\documentclass{bmvc2k}

\usepackage[nolist]{acronym}
\begin{acronym}[PHOENIX14\textbf{T} ]

\acrodefplural{rnn}[RNNs]{Recurrent Neural Networks}
\acrodefplural{cnn}[CNNs]{Convolutional Neural Networks}
\acrodefplural{hmm}[HMMs]{Hidden Markov Models}
\acrodefplural{gru}[GRUs]{Gated Recurrent Units}
\acrodefplural{crf}[CRFs]{Conditional Random Fields}
\acrodefplural{gan}[GANs]{Generative Adversarial Networks}
\acrodefplural{gpu}[GPUs]{Graphic Processing Units}

\acrodefplural{mdn}[MDNs]{Mixture Density Networks}

\acro{btg}[BTG]{Bracketing Transduction Grammar}
\acro{bpe}[BPE]{Byte Pair Encoding}
\acro{bsl}[BSL]{British Sign Language}
\acro{bleu}[BLEU]{Bilingual Evaluation Understudy}
\acro{bobsl}[BOBSL]{BBC-Oxford British Sign Language}
\acro{blstm}[BLSTM]{Bidirectional Long Short-Term Memory}
\acro{bslcpt}[BSLCP\textbf{T}]{BSL Corpus \textbf{T}}
\acro{cnn}[CNN]{Convolutional Neural Network}
\acro{crf}[CRF]{Conditional Random Field}
\acro{cslr}[CSLR]{Continuous Sign Language Recognition}
\acro{ctc}[CTC]{Connectionist Temporal Classification}
\acro{c4a}[C4A]{Content4All}
\acro{dl}[DL]{Deep Learning}
\acro{dgs}[DGS]{German Sign Language - Deutsche Gebärdensprache}
\acro{dsgs}[DSGS]{Swiss German Sign Language - Deutschschweizer Geb\"ardensprache}
\acro{dtw}[DTW]{Dynamic Time Warping}
\acro{dtwmje}[DTW-MJE]{Dynamic Time Warping Mean Joint Error}
\acro{fc}[FC]{Fully Connected}
\acro{ff}[FF]{Feed Forward}
\acro{fps}[fps]{frames per second}
\acro{gan}[GAN]{Generative Adversarial Network}
\acro{gpu}[GPU]{Graphics Processing Unit}
\acro{gru}[GRU]{Gated Recurrent Unit}
\acro{gtpt}[G2PT]{Gloss-to-Pose Transformer}
\acro{gtp}[G2P]{Gloss-to-Pose}
\acro{gts}[G2S]{Gloss-to-Sign}
\acro{gs}[GS]{Gloss Selection}
\acro{gr}[GR]{Gloss Reordering}
\acro{gt}[GT]{ground truth}
\acro{hmm}[HMM]{Hidden Markov Model}
\acro{hpe}[HPE]{Hand Pose Enhancer}
\acro{hoh}[HOH]{Hard of Hearing}
\acro{hns}[HamNoSys]{Hamburg Notation System}

\acro{isl}[ISL]{Irish Sign Language}
\acro{lstm}[LSTM]{Long Short-Term Memory}
\acro{mha}[MHA]{Multi-Headed Attention}
\acro{mo}[MoCap]{Motion Capture}
\acro{mtc}[MTC]{Monocular Total Capture}
\acro{mse}[MSE]{Mean Squared Error}
\acro{mdn}[MDN]{Mixture Density Network}
\acro{mdgs}[mDGS]{Meine DGS Annotated}
\acro{mdgst}[mDGS\textbf{T}]{meineDGS\textbf{T}}
\acro{mdgsth}[mDGS\textbf{T}-\textbf{H}]{meineDGS\textbf{T}-\textbf{HARD}}
\acro{mdgste}[mDGS\textbf{T}-\textbf{E}]{meineDGS\textbf{T}-\textbf{EASY}}
\acro{mt}[MT]{Machine Translation}

\acro{nmt}[NMT]{Neural Machine Translation}
\acro{nlp}[NLP]{Natural Language Processing}
\acro{nar}[NAR]{Non-AutoRegressive}
\acro{nsvq}[NSVQ]{Noise Substitution Vector Quantization}
\acro{ph12}[PHOENIX12]{RWTH-PHOENIX-Weather-2012}
\acro{ph14}[PHOENIX14]{RWTH-PHOENIX-Weather-2014}
\acro{ph14t}[PHOENIX14\textbf{T}]{RWTH-PHOENIX-Weather-2014\textbf{T}}
\acro{pof}[POF]{Part Orientation Field}
\acro{pos}[POS]{Part Of Speech}
\acro{pt}[PT]{Progressive Transformer}
\acro{paf}[PAF]{Part Affinity Field}
\acro{pttt}[P2TT]{Pose-to-Text Transformer}
\acro{ptt}[P2T]{Pose-to-Text}
\acro{pts}[P2S]{Pose-to-Sign}
\acro{ptgtt}[P2G2T]{Pose-to-Gloss-to-Text}
\acro{pca}[PCA]{Principal Component Analysis}
\acro{relu}[RELU]{Rectified Linear Units}
\acro{rnn}[RNN]{Recurrent Neural Network}
\acro{rouge}[ROUGE]{Recall-Oriented Understudy for Gisting Evaluation}
\acro{vqvae}[VQ-VAE]{Vector Quantized Variational Autoencoders}
\acro{vq}[VQ]{Vector Quantisation}
\acro{vae}[VAE]{Variational Autoencoders}
\acro{sgd}[SGD]{Stochastic Gradient Descent}
\acro{sla}[SLA]{Sign Language Assessment}
\acro{slr}[SLR]{Sign Language Recognition}
\acro{slt}[SLT]{Sign Language Translation}
\acro{slp}[SLP]{Sign Language Production}
\acro{smt}[SMT]{Statistical Machine Translation}
\acro{slo}[SLO]{Spoken Language Order}
\acro{so}[SO]{Sign Language Order}
\acro{snr}[S\&R]{Select and Reorder}
\acro{sse}[SSE]{Sign Supported English}
\acro{stt}[S2T]{Sign-to-Text}
\acro{stgtt}[S2G2T]{Sign-to-Gloss-to-Text}
\acro{ttgt}[T2GT]{Text-to-Gloss Transformer}
\acro{ttpt}[T2PT]{Text-to-Pose Transformer}
\acro{ttp}[T2P]{Text-to-Pose}
\acro{ttg}[T2G]{Text-to-Gloss}
\acro{tth}[T2H]{Text-to-HamNoSys}
\acro{ttgth}[T2G2H]{Text-to-Gloss-to-HamNoSys}
\acro{ttgtp}[T2G2P]{Text-to-Gloss-to-Pose}
\acro{tts}[T2S]{Text-to-Sign}
\acro{ttsse}[T2SSE]{Text to Sign Supported English}
\acro{wer}[WER]{Word Error Rate}
\acro{wmt14}[WMT2014]{WMT2014 German-English}

\end{acronym}

\usepackage{float}
\usepackage{amsfonts}
\usepackage{amsmath}
\usepackage{booktabs}
\usepackage{multicol} 
\usepackage[capitalise]{cleveref}

\title{Sign Stitching: A Novel Approach to Sign Language Production}

\addauthor{Harry Walsh}{harry.walsh@surrey.ac.uk}{1}
\addauthor{Ben Saunders}{b.saunders@surrey.ac.uk}{1}
\addauthor{Richard Bowden}{r.bowden@surrey.ac.uk}{1}

\addinstitution{
 CVSSP\\
 University of Surrey\\
 Guildford, UK
}

\runninghead{H. Walsh, B. Saunders, R. Bowden}{Sign Stitching}

\begin{document}

\maketitle

\begin{abstract}

Sign Language Production (SLP) is a challenging task, given the limited resources available and the inherent diversity within sign data. As a result, previous works have suffered from the problem of regression to the mean, leading to under-articulated and incomprehensible signing.
In this paper, we propose using dictionary examples to create expressive sign language sequences. However, simply concatenating the signs would create robotic and unnatural sequences. Therefore, we present a 7-step approach to effectively stitch the signs together. 
First, by normalising each sign into a canonical pose, cropping and stitching we create a continuous sequence. Then by applying filtering in the frequency domain and resampling each sign we create cohesive natural sequences, that mimic the prosody found in the original data. 
We leverage the SignGAN model to map the output to a photo-realistic signer and present a complete Text-to-Sign (T2S) SLP pipeline. 
Our evaluation demonstrates the effectiveness of this approach, showcasing state-of-the-art performance across all datasets.
  
\end{abstract}

\section{Introduction}
\label{sec:intro}
\acf{slp} is an essential step in facilitating two-way communication between the Deaf and Hearing communities. Sign language is inherently multi-channelled, with channels performed asynchronously and categorised into manual (hands and body) and non-manual (facial, rhythm, stress and intonation) features. For sign language to be truly understandable, both manual and non-manual features must be present. 

Analogous to the tone and rhythm used in spoken language, signed language exhibits prosody. The natural rhythm, stress and intonation that signed languages use to convey information \cite{wilbur2009effects}. 

\begin{figure}[tb]
  \centering
  \includegraphics[height=3.2cm]{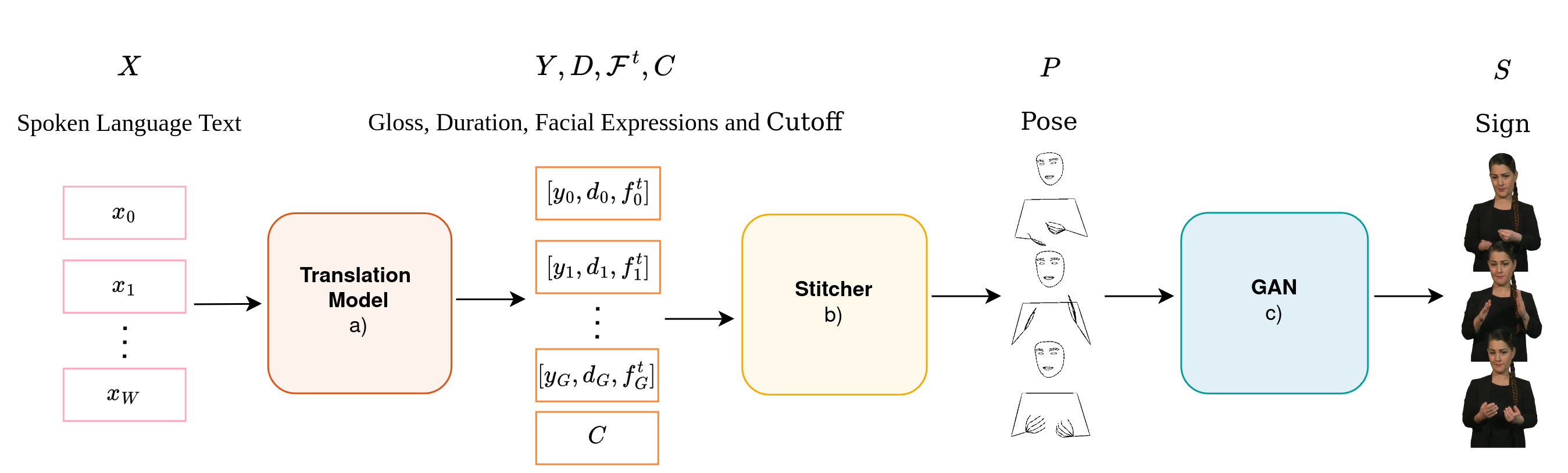}
  \caption{Overview of our approach. a) Spoken language to gloss, duration, facial expression and cutoff translation. b) Pose sequence generation. c) Photorealistic signer production.
  }
  \label{fig:overview}
\end{figure}

Sign language corpora containing linguistic annotation are limited due to the cost and time required to create such annotations \cite{dgscorpus_3}. Previous works have attempted to directly regress a sequence of poses from the spoken language or representations such as gloss \cite{saunders2020adversarial, saunders2021mixed, huang2021towards, hwang2023autoregressive, xie2022vector, xie2024g2p, stoll2020text2sign, tang2022gloss}. However, given that sign language is a low-resource language and the complexity is under-represented in small datasets, previous approaches have suffered from regression to the mean, resulting in under-articulated and incomprehensible signing. Additionally, previous works have implicitly modelled prosody, but due to the limited resources, it is often lost in production. 

In this paper, we propose a novel approach to \ac{slp} that effectively stitches together dictionary examples to create a meaningful, continuous sequence of sign language. By using isolated signs, we ensure the sequence remains expressive, overcoming previous shortcomings related to regression to the mean. However, each example lacks non-manual features, so we propose a \ac{nsvq} transformer architecture to learn a dictionary of facial expressions that can be added to each sign to create a realistic sequence. To the best of our knowledge, we are the first to explicitly model aspects of signed prosody in the context of \ac{slp}. By training a translation model to predict glosses, alongside a duration, facial expression and a cutoff value, we can modify the sequence to eliminate robotic and unnatural movements. Resampling each sign to the predicted duration allows us to alter the velocity associated with signing stress and rhythm \cite{wilbur1999stress}. Furthermore, by applying filtering in the frequency domain, we can adjust the trajectory of each sign to create softer signing, akin to how signers modify a sign to convey sentiment \cite{brentari1998prosodic, heloir2007qualitative, reilly1992affective}. Our approach demonstrates it is capable of modifying the stitched sequence to emulate aspects of prosody seen in the original continuous data. Evaluation of the produced sequence with back translation showcases state-of-the-art performance on all three datasets.

Furthermore, to conduct a realistic user evaluation we use SignGAN, a \ac{gan} capable of generating photo-realistic sign language videos from a sequence of poses \cite{saunders2020everybody}. Thus, we present a full \acf{tts} \ac{slp} pipeline that contains both manual and non-manual features. The user evaluation agrees that the approach outperforms the baseline method \cite{saunders2020progressive} and improves the realism of the signed sequence. An overview of the approach can be seen in \cref{fig:overview}.

\section{Related Work}
\label{sec:related_word}
\textbf{Sign language Translation:} For the last 30 years Computational \ac{slt} has been an active area of research \cite{tamura1988recognition}.  Initially focusing on isolated \ac{slr}
where a single instance of a sign was classified using a \ac{cnn} \cite{Lecun_Gradient_based_lrn}. Subsequent works extended to \ac{cslr}, which requires both the segmentation of a video into its constituent signs and their respective classification \cite{KOLLER2015108}. Later Camgoz et al. introduced the task of \ac{stt} \cite{camgoz2018neural} using neural networks, an extension of \ac{cslr} that requires the additional task of translation to spoken language. \ac{stt} performance was later improved using a Transformer \cite{vaswani2017attention}. 

Although there has been a lot of work since, the architecture has since become the standard when computing back-translation performance \cite{saunders2020adversarial, saunders2020progressive, saunders2021mixed}.
 
\textbf{Sign Language Production:} \ac{slp} is the reverse task to \ac{slt}, which aims to translate spoken sentences into continuous sign language. Early approaches to \ac{slp} used an animated avatar driven with either motion capture data or parameterised glosses \cite{bangham2000virtual, cox2002tessa, zwitserlood2004synthetic, efthimiou2012dicta, van2008virtual, elghoul2011websign}. These works all required expensive annotation systems, such as the \ac{hns} \cite{prillwitz1989hamnosys} or SigML \cite{kennaway2015avatar} and have shown to be unpopular with the Deaf community due to the robotic motion and under articulated signing \cite{rastgoo2021sign}. None of these approaches attempt to join the isolated signs effectively. Instead opting to play each sign in the sequence, with unnatural transitions in between. 

Early deep learning \ac{slp} approaches used \ac{nmt} and broke the task down into three steps,  \ac{ttg}, \ac{gtp} and \ac{pts} \cite{stoll2020text2sign}. Saunders et al. introduced the \ac{pt} \cite{saunders2020progressive}, a transformer architecture that synthesises poses directly from text. Although better results were achieved using gloss as an intermediate representation, the approach suffered from regression to the mean, caused by the lack of training data and the diversity of lexical variants. To reduce the problem, adversarial training and \ac{mdn} were applied \cite{saunders2020adversarial, saunders2021mixed} and since then a range of approaches have been proposed \cite{huang2021towards, hwang2023autoregressive, xie2022vector, xie2024g2p, tang2022gloss, saunders2021continuous}. However, visual inspection of the results shows that the approaches still suffer from regression to the mean, and as a result, they fail to effectively convey the translation. Here we propose a method to effectively join isolated signs, which means the produced sequences are guaranteed to be expressive and do not suffer from regression to the mean.

Furthermore, preliminary experiments reveal that each sign language sequence contains a distinct range of frequencies correlated to the signer's style. Fast motions contain high frequencies, while soft, slow signing involves lower frequencies, typically within the range of 1 to 25 Hz. To emulate this characteristic we filter the produced sequences in the frequency domain using a low-pass Butterworth filter \cite{butterworth1930theory}. This ensures the movements are stylistically cohesive. In addition, by adjusting the duration of each sign, we recreate the prosody observed in the original data.

\section{Methodology}
\label{sec:methodology}
\ac{slp} aims to facilitate the continuous translation from spoken to signed languages by converting a source spoken language sequence,  \(X = (x_{1},x_{2},...,x_{W})\) with W words into a video of photo-realistic sign, denoted as \(V = (v_{1}, v_{2}, \ldots,v_{U})\) with U frames. To accomplish this the approach uses two intermediate representations, following \cref{fig:overview} from left to right. First, the spoken language is translated to a sequence of glosses, \(Y = (y_{1},y_{2}, \ldots,y_{G})\), face tokens, \(\mathcal{F}^{t} = (f^{t}_{1}, f^{t}_{2}, \ldots, f^{t}_{G})\) and duration's, \(D = (d_{1},d_{2}, \ldots,d_{G})\), all with length \(G\). Additionally, for each sequence, we predict a low pass cutoff, \(\mathcal{C}\), which we use to filter the movements (\cref{fig:overview}.a). Each gloss and facial expression is stitched together using these parameters, to produce a continuous sequence of poses,  denoted as \(P = (p_{1}, p_{2},...,p_{U})\) with U frames (\cref{fig:overview}.b). Finally, we use the pose sequence to condition the SignGAN module allowing us to produce a photo-realistic signer. Next, we provide a detailed explanation of each step in our pipeline, following the order illustrated in \cref{fig:overview} from left to right. We then elaborate on the process of generating the cutoff frequencies and the dictionary of facial expressions.

\subsection{Translation Model}
\label{sec:methodology_t2g}
Given a spoken language sequence \(X = (x_{1},x_{2},...,x_{W})\), our goal is to generate a corresponding sequence of glosses \(Y = (y_{1},y_{2},...,y_{G})\). We design the transformer with four output layers, enabling the model to predict the corresponding duration (in frames) and facial expression for each gloss, plus a low-pass filter cutoff in Hz for each sequence. Thus the model learns the conditional probability \(p(Y, D, \mathcal{F}^{t}, \mathcal{C}|X)\).

\begin{figure}[tbh]
  \centering
  \includegraphics[height=3.2cm]{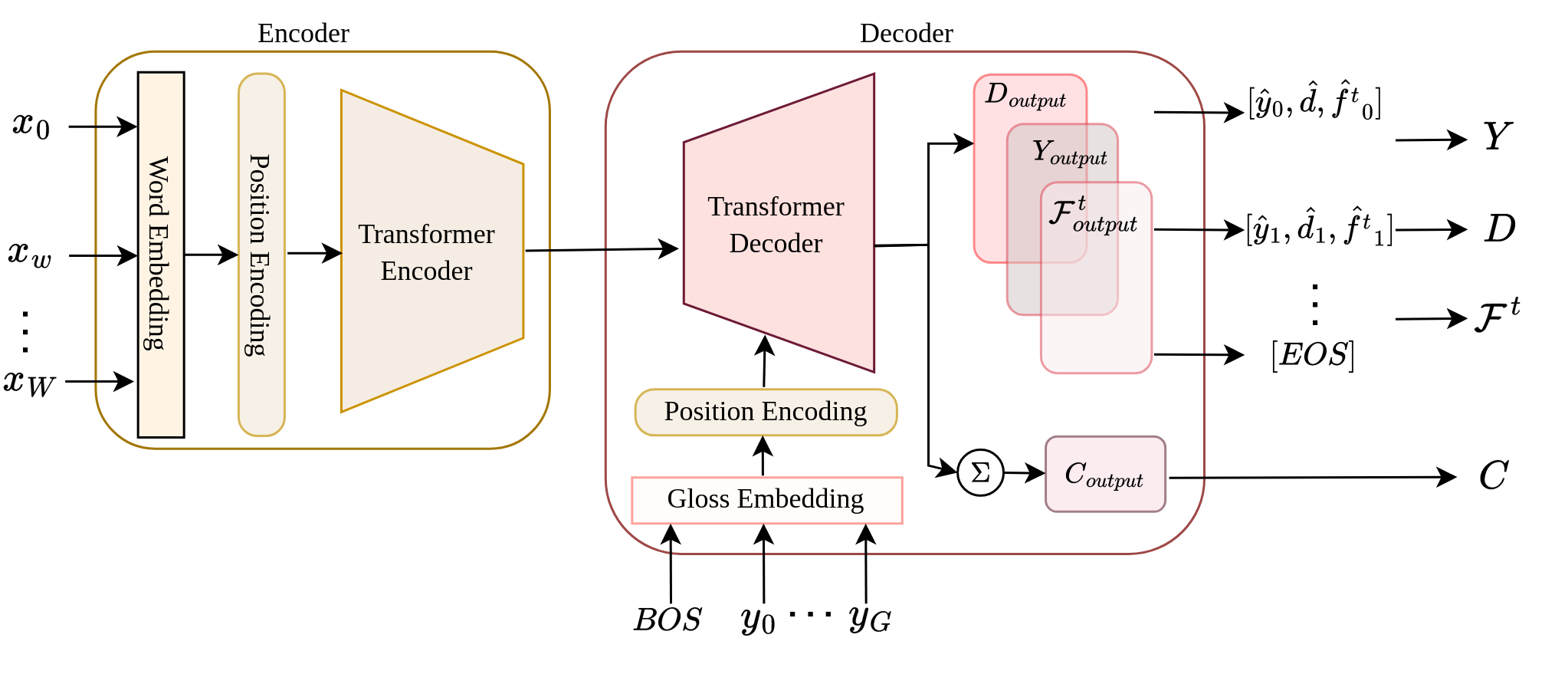}
  \caption{An overview of the Translation module.
  }
  \label{fig:translation}
\end{figure}

The model is an encoder-decoder transformer with \ac{mha}. The spoken language and gloss sequences are tokenized at the word level, and the embedding for a sequence is generated using a token embedding layer. Following the embedding layer, we add sine and cosine positional encoding.

The encoder learns to generate a contextualized representation for each token in the sequence. This representation is then fed into the decoder, which consists of multiple layers of self and cross \ac{mha} along with feedforward layers, and residual connections. The gloss, facial expression and duration predictions are obtained by passing the decoder output through their respective output layers. To obtain the cutoff prediction, we pool the decoder embedding across each time step and pass the output through a linear layer. The model is trained end-to-end with the following loss function; 
\begin{equation}
    L_{total} = \lambda_{y}\sum_{i=1}^{Y} \hat{y}_i \log(y_i) + \lambda_{f}\sum_{i=1}^{\mathcal{F}} \hat{F}_i \log(F_i) + \lambda_{d}\frac{1}{n} \sum_{i=1}^{n} (d_i - \hat{d}_i)^2 + \lambda_{C}(C - \hat{C})^2
\end{equation}
Each component is scaled by a factor, \(\lambda_{y}\), \(\lambda_{d}\), \(\lambda_{f}\) and \(\lambda_{C}\) before being combined to give the total loss, \(L_{total}\). The predictions from this model are passed to the stitching module to generate a pose sequence.

\subsection{Stitching}
\label{sec:methodology_stitch}

\begin{figure}[htb!]
  \centering
  \includegraphics[height=3.2cm]{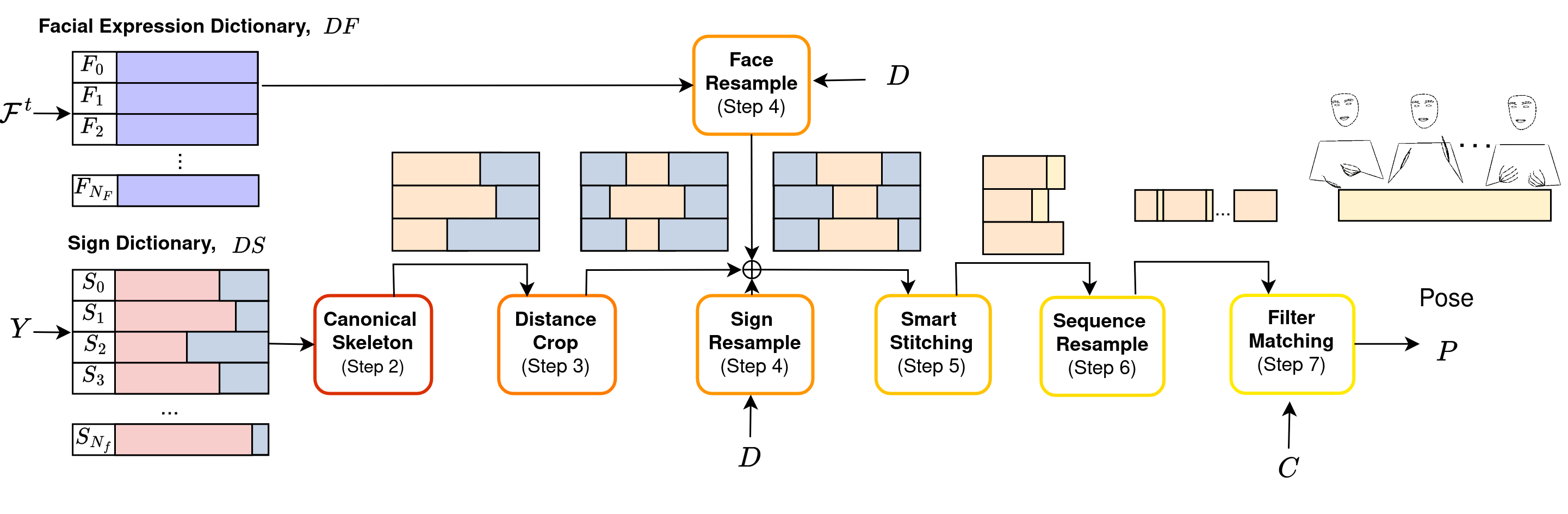}
  \caption{An overview of the stitching module.
  }
  \label{fig:stitcher}
\end{figure}

For each dataset, we collect an isolated instance of each gloss in our target vocabulary. For each sign, we extract Mediapipe skeletons \cite{lugaresi2019mediapipe} and run an additional optimization to uplift the 2D skeletons to 3D \cite{10193629}. The optimisation uses forward kinematics and a neural network to solve for joint angles, \(J_{a}\). We choose to store our dictionary as joint angles, as this allows us to apply a canonical skeleton. This ensures the stitched sequence is consistent even if the original signers have different bone lengths. We define a dictionary of, \(N_{s}\), signs as \(DS = [S_{1}, S_{2},..., S_{N_s}]\) where each sign in the dictionary consists of a sequence of angles, such that  \(S_{i} = (a_{1}, a_{2},..., a_{U_{s}})\) and \(a_{i} \in \mathbb{R}^{J_{a}}\), where \(U_{s}\) is the duration in frames. In addition we define a learnt dictionary of, \(N_{f}\), facial expressions as \(DF = [F_{1}, F_{2},..., F_{N_f}]\), where \(F_{i} \in \mathbb{R}^{U_{f} \times J \times D}\). \cref{fig:stitcher} illustrates our seven-step stitching pipeline, we now detail each step.

\textbf{Step 1}) Given a list of glosses, \(Y\), we select the corresponding signs in our dictionary. If a gloss is absent from the dictionary, we initially lemmatize and format the gloss. If still, we are unable to find a match in the dictionary, we apply a word embedding model and compute the cosine similarity with all words in the dictionary. We then select the closest sign as the substitute. Such that;  
\begin{align}
 j_{sub} = \arg\max_{j}( \frac{\sum_{j=1}^{N_{s}} \varepsilon(y_q) \cdot \varepsilon(DS^{y}_j)}{\sqrt{\varepsilon(y_q)^2} \cdot \sqrt{\sum_{j=1}^{N_{s}} \varepsilon(DS^{y}_j)^2}}) && S = DS[{j_{sub}}]
\end{align}
Here \(\varepsilon()\) represents the word embedding model, \(y_q\) is the query gloss and \(DS^{y}\) is the dictionary's corresponding gloss tags. We find word embeddings capture the meaning of words, enabling substitutions such as replacing RUHRGEBIET (RUHR AREA) with LANDSCHAFT (LANDSCAPE). 
Simultaneously, in this step, we select the corresponding facial expressions, \(F\), from the dictionary, \(DF\), given the predicted face tokens, \(\mathcal{F}^{t}\).   

\textbf{Step 2}) The selected signs are converted from angles into a 3D canonical pose. We normalise the rotation of the signer, such that the midpoint of the hips is located at the origin and the shoulders are fixed on the y plane. This ensures the skeleton is consistent across all the signs. Consequently, we convert from a sequence of angles \(S_{n}\in \mathbb{R}^{U_{\mathbf{s}}\times J_{a}}\) to a sequence of poses, \(P = (p_{1}, p_{2},...,p_{U_{s}})\) with the same number of frames, \(U_{s}\). Each pose, \(p_u\), is represented in \(D\)-dimensional space and consists of \(J\) joints, denoted as $p_{u} \in \mathbb{R}^{J \times D}$.

\textbf{Step 3}) The dictionary signs often start and end from a rest pose. Therefore, to avoid unnatural transitions we cropped the beginning and end of each sign. For this, we track the keypoint \(T\) corresponding to the wrist of the signer's dominant hand and measure the distance travelled.

Thus, for each dictionary sign we create a sequence, \(P^{\Delta} = (p^{\Delta}_2, p^{\Delta}_3, \ldots, p^{\Delta}_{U_{s}})\) \(n \in {2, 3, \ldots, U_{s}}\), representing the distance travelled for a dictionary sign. We remove the beginning frames once the sign has moved by a threshold, \(\alpha_{crop}\). The crop index is given by:
\begin{equation}
\text{index}_{\text{start}} = \arg\max_u \left( \sum_{u=1}^{U_{s}} P^{\Delta}_u - \alpha_{crop} \cdot \sum_{u=1}^{\max(U_{s})} P^{\Delta}_u \right)
\end{equation}
To crop the end, we reverse the order of frames and repeat the process. 

\textbf{Step 4}) As detailed above, we predict the duration of each gloss. Here, we utilize the duration to resample the length of each sign and facial expression, emulating the natural rhythm in the original data. This process involves upsampling or downsampling the sign using linear interpolation. Once the facial expression and sign are resampled to the same length, we shift and rotate the face onto the body creating the complete skeleton. 

\textbf{Step 5}) Having created a list of signs in a canonical pose, cropped, and resampled to match the original data, the next step involves joining these signs into a single sequence using the smart stitching module. The objective is to achieve a natural transition between the end of one sign and the start of the next. To accomplish this, we track the dominant hand of the signer and calculate the distance, \(\Delta\), between the end of the first sign and the start of the second sign. Then we can determine the required number of frames, \(U_{stitch}\), needed to create a smooth transition. Such that: 
\begin{equation}
    U_{stitch} = \arg\min_u(V_{min} < \frac{\text{fps} * \Delta}{u} < V_{max})
\end{equation}
Where \(V_{max}\) and \(V_{min}\) are the start and end velocities of the two signs. This calculation ensures that the signer's velocity is bounded by the end velocity of sign one and the initial velocity of sign two. 
In cases where multiple solutions exist, we select \(U_{stitch}\) that minimizes the standard deviation between the start and end velocity. Following this, we employ linear interpolation to generate the missing frames. 

\textbf{Step 6}) We concatenate the signs and the stitched frames to form a single sequence. We then sum all the predicted durations and resample the sequence to match the ground truth. 

\textbf{Step 7}) Finally, we apply a low-pass Butterworth filter to each keypoint over time \cite{butterworth1930theory}. The predicted cutoff value determines which frequencies are removed, and corresponds to the -3dB attenuation point. This step aims to enhance the stylistic cohesiveness of the sequence by smoothing out any sharp, quick movements not present in the original sequence. The transfer function can be formulated as;

\begin{equation}
    H(z) = \left((1 + \left(\frac{z}{{e^{j \cdot \omega_c}}}\right)^{2N}\right)^{-1}
    \label{eq:filter}
\end{equation}

Here we apply a 4th order filter thus, N is 4, and the angular cutoff frequency is given by \( \omega_c = 2\pi C\). z corresponds to the z-transform of the pose sequence. Applying the bilinear transform to \cref{eq:filter} gives the discrete formula that we apply. This process generates natural pose sequences, that remain expressive and are stylistically cohesive. Next, we map these poses to a photo-realistic signer. 

\subsection{SignGAN}
\label{sec:methodology_gan}

Skeleton outputs have been shown to reduce Deaf comprehension compared to a photo-realistic signer \cite{ventura2020can}. Therefore, to gain valuable feedback from the Deaf community we train a SignGan model~\cite{saunders2020everybody}.  Given a pose sequence, \(P = (p_{1}, p_{2}, \ldots, p_{U})\), generated by our stitching approach the model aims to generate the corresponding video of sign language, \(V = (v_{1}, v_{2}, \ldots,v_{U})\) with \(U\) frames. 

\subsection{Facial Expression Generation}
To be truly understandable and accepted by the Deaf community non-manual features must be present in the final output. Here we are using a discrete sequence-to-sequence model to generate the translation. Therefore, we must learn a discrete vocabulary of facial expressions, \(DF\), that can be added to the isolated signs. We design a transformer base \ac{nsvq} architecture to learn a spatial-temporal dictionary of facial expressions. Using the ground truth gloss timing information, we extract the corresponding face mesh sequence, denoted as \(F\). We then resample each sequence to a constant length, \(U_{f}\) and scale it to be a constant size. Signers in the dataset are often looking off center, therefore we normalize the average direction of the face so that it is looking directly forward.
Similar to \cref{fig:translation} (Encoder) we add positional encoding and then embed the sequence using a single linear layer. After the embedding is passed through the transformer encoder to the codebook. The \ac{nsvq} codebook learns a set of \(N_{f}\) embeddings \cite{vali2022nsvq}. We denote the embedded face sequence and therefore each codebook entry as \(F^{z}_{i} \in \mathbb{R}^{U_{f} \times H }\), where \(H\) is the embedding dimension. Each input is mapped to one codebook entry, the difference between the selected codebook entry and the input is then simulated using a normally distributed noise source. A product of the simulated noise and the encoder output is then passed to the decoder. We use the counter decoding technique from the \ac{pt} \cite{saunders2020progressive}, to drive the decoder. The decoder learns to reconstruct the original face sequence and the input counter values. Thus, the model is trained end-to-end with the following loss function;
\begin{equation}
L_{Face} = \dfrac{1}{U_{f}} \sum_{u=1}^{U_{f}}((f_{u} - \hat{f_{u}})^2 + \lambda_{CN}(c_{u} - \hat{c_{u}})^2)
\end{equation}
Where \(\lambda_{CN}\) is a scaling factor and \(c\) is the counter value. Once fully trained we pass each codebook embedding, \(F^{z}_{i}\), through the decoder to give the learnt dictionary of facial expressions in Euclidean space, \(DF = [F_{1}, F_{2},..., F_{N_f}]\).

\subsection{Cutoff Generation}

To generate the ground truth cutoffs used in training, we once again apply our stitching approach. For each sequence in the ground truth data, \(P\), we produce the equivalent stitched sequence, \(P_{stitch}\). We then apply a low-pass filter to \(P_{stitch}\) within the range of 1 to 25 Hz and measure the intersection and set difference of the frequencies, denoted as \((FT(P) \cap FT(P_{stitch}))\) and \((FT(P) \setminus FT(P_{stitch}))\), where the Fourier transform is represented as \(FT()\). Subsequently, we fit a parametric spline curve to the intersection and set difference. To determine the cutoff we find the frequency that maximises the intersection while minimising the set difference. This provides the cutoff frequency for that sequence.
We opt to use this method over just analyzing the frequency in the original sequence as we do not have an ideal filter. Thus, the Butterworth filter has unintended effects on frequencies below the cutoff.


\section{Experiments}
\label{sec:experiments}
\subsection{Implementation Details}

We apply the approach to three datasets, the Public Corpus of German Sign Language, 3rd release, the \ac{mdgs} dataset \cite{dgscorpus_3}, \ac{ph14t} \cite{camgoz2018neural} and the \ac{bslcpt} \cite{schembri2008british}. To evaluate our approach we employ a \ac{cslr} model (Sign Language Transformers \cite{camgoz2020sign}) to conduct back-translation, the same as \cite{huang2021towards, saunders2020adversarial, saunders2021mixed, xie2022vector}. BLEU \cite{papineni2002bleu}, Rouge \cite{lin2004rouge}, and chrF \cite{popovic-2015-chrf} scores are computed between the predicted text and the ground truth. Finally, to evaluate the pose we employ \ac{dtwmje}. In the following experiment, we test two different dictionaries: 1) collected from isolated examples, and 2) a dictionary created from continuous data. Noted as Isolated and continuous in the following tables. Further information about the datasets, dictionaries and model implementation can be found in the supplementary material.

\subsection{Quantitative Evaluation}
\textbf{Text-to-Gloss Translation Results:} We start by evaluating the \ac{ttg} translation performance described in \cref{sec:methodology_t2g}. \cref{tab:t2g} shows the performance on all three datasets. We suggest that the difficulty of a dataset is proportional to the vocabulary and the total number of sequences used in training. We find the best performance on \ac{ph14t} data which has the highest number of sequences per token, achieving 18.11 BLEU-4. In comparison to previous works, we find by having the model predict duration, face and cutoff we can achieve higher BLEU-1 scores, but at the cost of a lower BLEU-4 in comparison to \cite{saunders2020progressive}.  
On the more challenging \ac{mdgs} dataset we find a considerably lower BLUE-4 score due to the larger domain of discourse. The \ac{bslcpt} has a smaller domain of discourse in comparison to \ac{mdgs} but has the fewest sequences per token. Thus, understandably we only achieve a BLEU-4 of 1.67 on the test set. Overall we find the model to be performing as expected.

\begin{table*}[h!]
\centering
\resizebox{\linewidth}{!}{%

\begin{tabular}{r|cccccc|cccccc} 
\toprule
\multicolumn{1}{l}{}      & \multicolumn{6}{c}{TEST SET}       & \multicolumn{6}{c}{DEV SET}  \\
\multicolumn{1}{c|}{Dataset:} & BLEU-1   & BLEU-2 & BLEU-3 & BLEU-4 & chrF & ROUGE & BLEU-1 & BLEU-2  & BLEU-3 & BLEU-4 & chrF & ROUGE \\
\midrule
BSLCP\textbf{T}     & 26.02    & 11.19   & 4.15   & 1.67   & 23.54 & 25.06 & 26.88  & 11.40    & 4.72   & 1.28   & 23.59 & 26.96 \\
mDGS                & 30.13         & 13.04       & 5.45       & 2.36       & 29.24     & 31.43      & 29.72       & 12.46        & 4.89       & 1.87       &  28.90    & 30.90      \\
\hline
PHOENIX14\textbf{T} & 55.48         & 36.54       & 25.18       & 18.11       & 49.30     & 53.83      & 56.55       & 37.32        & 25.85       & 18.74       & 48.91     & 54.81      \\
PHOENIX14\textbf{T} \cite{saunders2020progressive} & 55.18  & 37.10 & 26.24 & 19.10 & - & 54.55 & 55.65 & 38.21 & 27.36 & 20.23 & -  & 55.41 \\ 
PHOENIX14\textbf{T} \cite{stoll2018sign} & 50.67 & 32.25 & 21.54 & 15.26 & -  & 48.10 & 50.15 & 32.47 & 22.30 & 16.34 & - & 48.42 \\
\bottomrule
\end{tabular}
}
\caption{\label{tab:t2g}The results of translating from Text-to-Gloss on the BSL Corpus \textbf{T}, RWTH-PHOENIX-Weather-2014\textbf{T} and Meine DGS Annotated dataset.}
\end{table*}

\textbf{Text-to-Pose Translation Results:} Note in the following experiments the back-translation model's performance (shown as GT top row of \cref{tab:BSLCP}, \ref{tab:MDGS} and \ref{tab:PHIX}) should be considered the upper limit of performance. 
In this section, we evaluate the \ac{ttp} performance using back translation. To allow for a comparison we train two versions of the \ac{pt} with the setting presented in \cite{saunders2020progressive}. \ac{pt} is the standard architecture, while \ac{pt} + GN is trained with Gaussian Noise added to the input. In line with the original paper, we find Gaussian Noise improves the performance, however, our approach still outperforms both models except on \ac{dtwmje}. As discussed previously, other works suffer from regression to the mean caused by the models attempting to minimise their loss function and thus, are incentivised to predict a mean pose. This metric fails to evaluate the content of the sequence, but the higher score does indicate our model is expressive as it is producing sequences further from the mean.
For back-translation, we outperform the \ac{pt} on all metrics. Showing significant improvements in BLEU-1 score of 98\% and 269\% on the \ac{mdgs} and \ac{bslcpt} dev set (comparing \ac{pt} + GN and Stitcher (continuous), \cref{tab:BSLCP} and \ref{tab:MDGS}).

\begin{table*}
\centering

\resizebox{\linewidth}{!}{%

\begin{tabular}{r|ccccccc|ccccccc} 
\toprule
\multicolumn{1}{l}{BSLCP\textbf{T}}      & \multicolumn{7}{c}{TEST SET}       & \multicolumn{7}{c}{DEV SET}  \\
\multicolumn{1}{c|}{Approach:} & DTW-MJE & BLEU-1 & BLEU-2 & BLEU-3 & BLEU-4 & chrF & ROUGE & DTW-MJE & BLEU-1 & BLEU-2 & BLEU-3 & BLEU-4 & chrF & ROUGE   \\ 
\midrule
GT                                             & 0.000    & 17.3       & 3.96       & 1.37       & 0.54       & 13.00     & 21.76     & 0.000   & 17.32       & 3.71       & 1.08       & 0.39       & 13.04     & 21.89      \\
\midrule
PT  \cite{saunders2020progressive}             & 0.288    & 4.40   & 0.65       & 0.18       & 0.00   & 5.80 & 8.22     & 0.292        & 4.00       & 0.61       & 0.10       & 0.00       & 5.69     & 8.02      \\
PT + GN  \cite{saunders2020progressive}        & 0.267         & 4.96       & 0.55       & 0.13       & 0.00       & 6.38  & 8.82   & 0.258     & 4.47        & 0.63       & 0.09       & 0.00        & 6.14     &  8.89     \\
\midrule
Stitcher (Isolated)                                & 0.588         & 16.37       & 2.86       & 0.75       & 0.28       & 14.07     & \textbf{20.84}     & 0.592        & 16.39       & 2.82       & 0.58       & 0.00       & 13.9     &  19.55     \\
Stitcher (continuous)                                & 0.575     & \textbf{16.99}       & \textbf{3.65}       & \textbf{1.03}       & \textbf{0.41}       & \textbf{14.32}     & 20.65     & 0.573       & \textbf{16.52}       & \textbf{3.19}       & \textbf{0.73}       & 0.00       & \textbf{14.34}     & \textbf{20.53}      \\
\bottomrule
\end{tabular}

}
\caption{\label{tab:BSLCP}The results of translating from Text-to-Pose on the BSL Corpus \textbf{T} dataset.}
\end{table*}

Deep learning models exhibit a bias toward the data they were trained on and often show poor out-of-domain performance. Unsurprisingly, the performance improves when using the continuous dictionary. We find only a small increase in BLEU-1 of 0.13 on the \ac{bslcpt} dev set (\cref{tab:BSLCP}), most likely due to the isolated dictionary containing the lexical variants found in the original data. Whereas we see a larger increase on the \ac{mdgs} dataset (\cref{tab:MDGS}).

\begin{table*}
\centering
\resizebox{\linewidth}{!}{%

\begin{tabular}{r|ccccccc|ccccccc} 
\toprule
\multicolumn{1}{l}{mDGS}      & \multicolumn{7}{c}{TEST SET}               & \multicolumn{7}{c}{DEV SET}          \\
\multicolumn{1}{c|}{Approach:} & DTW-MJE & BLEU-1 & BLEU-2 & BLEU-3 & BLEU-4 & chrF & ROUGE & DTW-MJE & BLEU-1 & BLEU-2 & BLEU-3 & BLEU-4 & chrF & ROUGE \\ 
\midrule
GT                                       & 0.000    & 20.87  & 5.60   & 1.89   & 0.80   & 17.56   & 23.78 & 0.000   & 20.75  & 5.43   & 1.81   & 0.76   & 17.63     & 23.41 \\
\midrule
PT  \cite{saunders2020progressive}       & 0.229   & 6.11   & 0.94   & 0.21   & 0.05   &  8.07    & 8.36  & 0.228  & 6.22   & 0.98   & 0.17   & 0.00   &  8.23    & 8.44  \\
PT + GN  \cite{saunders2020progressive}  & 0.2245   & 7.18   & 1.48   & 0.40   & 0.01   &  8.46    & 8.38  & 0.2241  & 9.22   & 1.63   & 0.38   & 0.01   &  8.94    & 8.57  \\
\midrule
Stitcher (Isolated)   & 0.581         & 16.63       & 3.75       & 0.94       & 0.22       & 13.39     & 21.69      & 0.592  & 16.9        & 3.67       & 0.95       & 0.32       & 13.9     & \textbf{21.34}      \\
Stitcher (Continuous)   & 0.637    & \textbf{18.64}       & \textbf{4.17}       & \textbf{1.07}       & \textbf{0.39}       & \textbf{16.86}     & \textbf{21.80}     & 0.637  & \textbf{18.27}       & \textbf{4.07}      & \textbf{1.19}       & \textbf{0.43}       & \textbf{16.75}     & 21.25      \\
\bottomrule
\end{tabular}
}
\caption{\label{tab:MDGS}The results of translating from Text-to-Pose on the Meine DGS Annotated (mDGS) dataset.}
\end{table*}
\begin{table}
\centering
\resizebox{0.6\linewidth}{!}{%
\begin{tabular}{r|ccccccc} 
\toprule
\multicolumn{1}{l}{PHOENIX14\textbf{T}}      & \multicolumn{7}{c}{}          \\
\multicolumn{1}{c|}{Approach:} & DTW-MJE & BLEU-1 & BLEU-2 & BLEU-3 & BLEU-4 & chrF & ROUGE   \\ 
\midrule
 GT                                      & 0.000   & 32.41  & 20.19  & 14.41  & 11.32  & 33.84   & 32.96    \\
 \midrule
 PT \cite{saunders2020progressive}       & 0.197   & 6.27   & 3.33   & 2.14   & 1.59   &  14.52   & 9.50    \\
 PT + GN \cite{saunders2020progressive}  & 0.191   & 11.45  & 7.08   & 5.08   & 4.04   &  19.09   & 14.52   \\
 NAT-AT \cite{huang2021towards}          & 0.177   & 14.26  & 9.93   & 7.11   & 5.53   &  21.87   & 18.72   \\
 NAT-EA \cite{huang2021towards}          & 0.146   & 15.12  & 10.45  & 7.99   & 6.66   &  22.98   & 19.43   \\
 PoseVQ-MP \cite{xie2022vector}          & 0.146   & 15.43  & 10.69  & 8.26   & 6.98   &  -       & -       \\
 PoseVQ-DDM \cite{xie2022vector}         & 0.116   & 16.11  & 11.37  & \textbf{9.22}   & \textbf{7.50}   &  -       & -       \\
 \midrule
Stitching G2P (Isolated)                        & 0.593   & 21.47  & 8.79   & 4.25   & 2.49   & 23.74 & 20.32  \\
Stitching G2P (Continuous)                        & 0.587   & 23.58   & 12.31   & 8.05   & 5.95   & 28.85 & 24.84   \\
Stitching T2P (Isolated)                        & 0.594   & 22.78  & 9.68   & 5.17   & 3.12   & 24.27 & 21.30  \\
Stitching T2P (Continuous)                        & 0.572   & \textbf{25.14}  & \textbf{13.54}   & 8.98   & 6.67   & \textbf{29.5} & \textbf{26.49}   \\
\bottomrule
\end{tabular}
}
\caption{\label{tab:PHIX}The results of translating from Gloss-to-Pose (G2P) and Text-to-Pose (T2P) on the RWTH-PHOENIX-Weather-2014\textbf{T} test set.}
\end{table}

Previous work has primarily focused on \ac{gtp} translation, therefore to facilitate a meaningful comparison we present two versions of the model. First, a \ac{gtp} version, where we use the ground truth data and just apply the stitching module, and, secondly our \ac{ttp} approach (translation then stitching). Results for comparison are provided by \cite{xie2022vector}. Note that the previous approaches do not use a dictionary of signs and instead attempt to regress the pose directly from the spoken language. 
We find our approach outperforms previous work on the BLEU-1 to 2 scores increasing the score by 56\% and 19\%, respectively (comparing \cref{tab:PHIX}, row 7 and 11). We also find significant improvement in ROUGE and chrF metrics.  

Using a continuous dictionary we can outperform all models except for the \ac{vq} based approaches on BLEU-3 to 4. As the VQ model is learning sub-units of a gloss sequence we suggest this gives it an advantage on higher n-grams, as each token that is predicted can represent multiple signs.

\subsection{Qualitative Evaluation}

\textbf{Visual Outputs:} To demonstrate the approach's effectiveness, we present skeleton and video outputs for two sign languages (BSL and DGS)\footnote{\url{https://github.com/walsharry/Sign_Stitching_Demos}}. 
Furthermore, in the supplementary material, we share visualizations of the produced sequence.    

\textbf{Survey Results:} The survey presented a comparison to \ac{pt}, followed by an ablation of different components of the stitching approach.  
17\% of people surveyed were native Deaf signers, while 34\% were L2 signers or language learners. 87.5\% preferred our approach compared to the \ac{pt}, while the rest selected no preference. 100\% of people agreed that applying the filter improved the realism compared, while resampling was found to be less important, with 37.5\% selecting no preference between the resampled and normal sequence.

\section{Conclusion}
\label{sec:conclusion}
In this paper, we presented a novel approach to \ac{slp}. Previous works have suffered from the problem of regression to the mean and have mainly focused on manual features. Here we have overcome the problem by using a dictionary of expressive examples. The stitching effectively joins the signs together creating a natural continuous sequence and by clustering facial expressions into a vocabulary we can create a sequence that contains both manual and non-manual features. We eliminated unnatural transitions and enhanced the stylistic cohesiveness through the approach. As a result, we present state-of-the-art performance. Finally, the user evaluation agrees with the quantitative results, indicating our approach can produce realistic expressive Sign language.

\section{Acknowledgement}
We thank Adam Munder, Mariam Rahmani, and Abolfazl Ravanshad from OmniBridge, an Intel Venture. We also thank Thomas Hanke and the University of Hamburg for the use of the Meine DGS Annotated (mDGS) data.
This work was supported by Intel, the SNSF project ‘SMILE II’ (CRSII5 193686), the European Union’s Horizon2020 programme (‘EASIER’ grant agreement 101016982) and the Innosuisse IICT Flagship (PFFS-21-47). This work reflects only the author's view and the Commission is not responsible for any use that may be made of the information it contains.


\bibliography{egbib}
\end{document}